\documentclass[sigconf]{acmart}
\usepackage[ruled,vlined]{algorithm2e}
\usepackage{enumitem}
\usepackage[
    disable,
    colorinlistoftodos,
    prependcaption,
    textsize=tiny
]{todonotes}

\AtBeginDocument{%
  \providecommand\BibTeX{{%
    \normalfont B\kern-0.5em{\scshape i\kern-0.25em b}\kern-0.8em\TeX}}}

\setcopyright{acmcopyright}
\copyrightyear{2020}
\acmYear{2020}
\acmDOI{10.1145/1122445.1122456}

\acmConference[San Diego '20]{KDD 2020 Conference Workshop
ACM SIGKDD Conference on Knowledge Discovery and Data Mining
}{August 24, 2020}{San Diego, California USA}
\acmBooktitle{San Diego '20: KDD 2020 Conference Workshop
ACM SIGKDD Conference on Knowledge Discovery and Data Mining
  August 24, 2020, San Diego, California USA}
\acmPrice{15.00}
\acmISBN{978-1-4503-XXXX-X/18/06}

\begin{document}

\title{An LSTM approach to Forecast Migration using Google Trends}

\author{Nicolas Golenvaux}
\authornote{Both authors contributed equally to this research.}
\author{Pablo Gonzalez Alvarez}
\orcid{0000-0003-2120-8815}
\authornotemark[1]
\affiliation{%
  \institution{UCLouvain, Belgium}
  \streetaddress{Place Sainte Barbe 2 bte L5.02.01}
  \postcode{1348}
}

\author{Harold Silv\`{e}re Kiossou}
\affiliation{%
  \institution{UCLouvain, Belgium}
  \streetaddress{Place Sainte Barbe 2 bte L5.02.01}
  \postcode{1348}
}

\author{Pierre Schaus}
\affiliation{%
  \institution{UCLouvain, Belgium}
  \orcid{0000-0002-3153-8941}
  \streetaddress{Place Sainte Barbe 2 bte L5.02.01}
  \postcode{1348}
}
\email{pierre.schaus@uclouvain.be}






\renewcommand{\shortauthors}{Golenvaux and Gonzalez Alvarez, et al.}


\begin{abstract}
Being able to model and forecast international migration as precisely as possible is crucial for policymaking.
Recently Google Trends data in addition to other economic and demographic data have been shown to improve the forecasting quality of a gravity linear model for the one-year ahead forecasting.
In this work, we replace the linear model with a long short-term memory (LSTM) approach and compare it with two existing approaches: the linear gravity model and an artificial neural network (ANN) model.
Our LSTM approach combined with Google Trends data outperforms both these models on various metrics in the task of forecasting the one-year ahead incoming international migration to 35 Organization for Economic Co-operation and Development (OECD) countries: for example the root mean square error (RMSE) and the mean average error (MAE) have been divided by 5 and 4 on the test set.
This positive result demonstrates that machine learning techniques constitute a serious alternative over traditional approaches for studying migration mechanisms.
\end{abstract}






\maketitle

\section{Introduction}
\label{sec:intro}
Mobility has always been part of human history. In 2017, there were about 258 million international migrants worldwide, of which 150.3 million are migrant workers~\cite{undesaInternationalMigrantStock2019}. Modeling and forecasting human mobility is therefore important not only to help formulate effective governance strategies but also to deliver insight at scale to humanitarian responders and policymakers.
However, developing reliable methods able to forecast $T_{i,j}$, the number of people moving at the next time step from a given region $i$ to another region $j$ among $m$ origin regions and $n$ destination regions is extremely challenging due to the absence of, low frequency and long lags in recent migration data, especially for developing countries~\cite{askitasInternetDataSource2015,bohmeSearchingBetterLife2020}.

One way to mitigate this lack of timely data is the use of real-time geo-referenced data on the internet like the Global Database of Events, Language, and Tone (GDELT Project) or Google Trends. Both have been successfully used to make forecasting in various fields \cite{choiPredictingPresentGoogle2012,ahmedMultiScaleApproachDataDriven2016,google_flue}.
Recently, \citeauthor{bohmeSearchingBetterLife2020}~\shortcite{bohmeSearchingBetterLife2020} demonstrated that adding geo-referenced online search data to forecast migration flows yields better performance compared to only using common economic and demographic indices like gross domestic product (GDP), and population size.
The authors propose to forecast bilateral migration flows of the following year with a linear model relying on the Google Trends data captured the previous year.

In this work, we use the same data, but we replace the linear model by a recurrent neural network (LSTM~\cite{hochreiterLongShorttermMemory1997}) that can consider the whole history to make forecasts.
We demonstrate that the forecasting quality can be drastically improved by capturing better complex migration dynamics~\cite{masucciGravityRadiationModels2013} and complex interactions between the many features.

The outline of our work is the following.
We first introduce the related work in section~\ref{sec:relatedwork}.
In order to make this article more self-contained, we explain in section \ref{sec:background} how the Google Trends features are extracted in \citeauthor{bohmeSearchingBetterLife2020}~\shortcite{bohmeSearchingBetterLife2020} and also briefly introduce recurrent neural networks.
We then describe our recurrent neural network approach in section \ref{sec:approach}.
Finally, our approach is evaluated and compared with the previous approaches in section \ref{sec:results_and_discussion}.

\section{Related Work}
\label{sec:relatedwork}

In traditional models, the problem of forecasting $T_{i,j}$ (i.e., to find $\hat{T}_{i,j}$) is usually divided into two sub-problems: (a) forecast $G_i$ the number of people leaving a region $i$ (also known as the production function); and (b) forecast $P_{i,j}$ the probability of a movement from $i$ to $j$. Thus we have $\hat{T}_{i,j} = G_i P_{i,j}$.

\begin{table*}
    \centering
    \caption{The input features used for the different models. Each feature spans from 2004 to 2014 for a pair of origin-destination country. Refer to subsection~\ref{par:gti} for a detailed explanation of the Google Trends Index (GTI).}
    \label{tab:features}
    \begin{tabular}{@{}lll@{}}
        \toprule
        \textbf{Input features$_{i, j, t}$} & & \textbf{Description} \\ 
        \midrule
        $GDP_{i,t}$ & & Gross Domestic Product for origin country $i$ during the year $t$ \\
        $GDP_{j,t}$ & & Gross Domestic Product for destination country $j$ during the year $t$ \\
        $pop_{i,t}$ & & Population size for origin country $i$ during year $t$\\
        $pop_{j,t}$ & & Population size for destination country $j$ during year $t$ \\
        $fixed_{i}$ & & Origin country $i$ fixed effects, encoded as a one-hot vector \\
        $fixed_{j}$ & & Destination country $j$ fixed effects, encoded as a one-hot vector \\
        $fixed_{t}$ & & Year $t$ fixed effects, encoded as a one-hot vector \\
        $GTIbi_{i,j,t}$ & & Bilateral GTI for a pair origin country $i$ and destination country $j$ during a year $t$\\
        $GTIuni_{i,t} \times GTIdest_{i,j,t}$ & & Unilateral and destination GTI for an origin country $i$, a destination country $j$ during a year $t$ \\
        $T_{i,j,t}$ & & Current year migration flow from country $i$ to country $j$  \\
        \bottomrule
    \end{tabular}
\end{table*}

There are two conventional models: (a) the gravity model; and (b) the radiation model.
Gravity models, inspired by Newton's law, evaluate the probability of a movement between two regions to be proportional to the population size of the two regions $i$ and $j$, and inversely proportional to the distance between them \cite{andersonGravityModel2011,letouzeRevisitingMigrationdevelopmentNexus2009,pootGravityModelMigration2016b}. In radiation models, inspired by diffusion dynamics, a movement is emitted from a region $i$ and has a certain probability of being absorbed by a neighboring region $j$. The subtlety here is that this probability is dependent on the population of origin, the population of the destination, and the population inside a circle centered in $i$ with a radius equal to the distance from $i$ to $j$ \cite{siminiUniversalModelMobility2012}. Gravity is usually better to capture short distance mobility behaviors, while radiation is usually better to capture long-distance mobility behaviors \cite{masucciGravityRadiationModels2013}.

With machine learning (ML) models, the approach is quite different as the goal is to directly forecast $\hat{T}_{i,j} = f(features)$ from a set of features\footnote{Notice that you could approach the problem the same way as with traditional models but it is not a common practice.}. To the best of our knowledge, \cite{robinsonMachineLearningApproach2018} is the first attempt to use ML to forecast human migration. The authors use two ML techniques: (a) \emph{"extreme" gradient boosting regression (XGBoost)} model; and (b) deep learning based \emph{artificial neural network (ANN)} model.
Similarly to us, this approach also attempts to directly forecast $\hat{T}_{i,j}$ from the set of features without requiring any production function.
But this approach exhibits two important differences with our approach:
a) it uses traditional features for their forecasting model, which is composed of geographical and econometric properties such as the inter-country distance, and the median household income; and b) it does not capture the dynamic aspect since the forecast only relies on the previous time-step set of features.

More recently, \citeauthor{bohmeSearchingBetterLife2020}~\shortcite{bohmeSearchingBetterLife2020}, use the Google Trends Index (GTI) of a set of keywords related to migration (e.g., visa, migrant, work) as a new feature set to make migration forecasts.
\citeauthor{bohmeSearchingBetterLife2020}~\shortcite{bohmeSearchingBetterLife2020} rely on a bilateral gravity model to forecast the total number of migrant leaving a country of origin towards any of the OECD’s destination countries during a specific year. The gravity models are estimated by a linear regression. Our approach uses the same input data and thus also relies on the GTI data. But instead of a linear least-squares estimation model, we propose a recurrent neural network (LSTM) that uses the complete set of historical features rather than only the ones coming from the previous time step.

\section{Background}
\label{sec:background}
\begin{figure*}
    \centering     \includegraphics[width=1.0\textwidth]{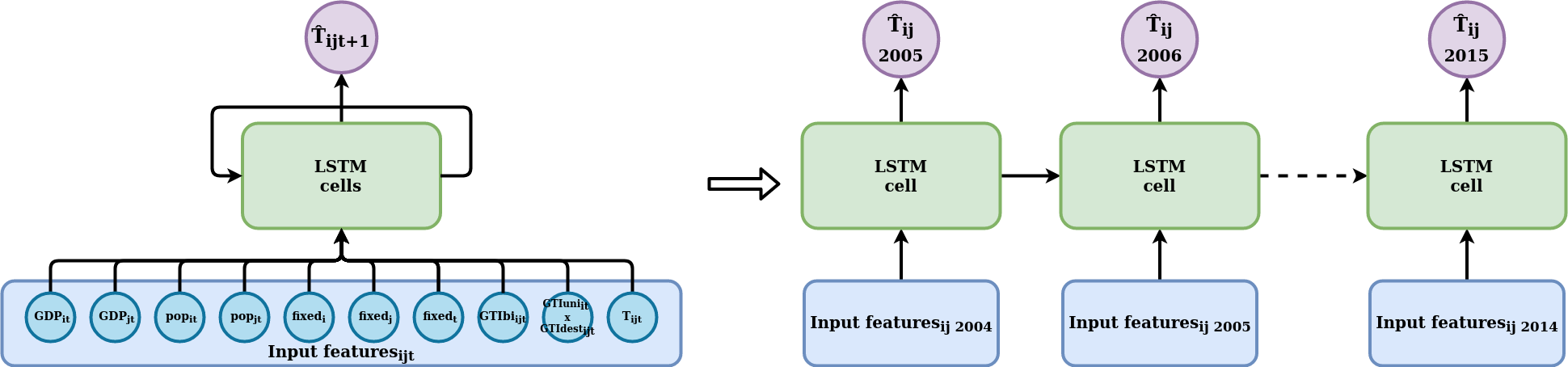}     \caption{An unfolded gated-RNN with LSTM cells. The left-side corresponds to the folded RNN, while the right-side to the unfolded RNN.}
\label{fig:unfolded_lstm} \end{figure*}

In this section, first, we describe the data used for learning and forecasting the migration, giving more details about the Google Trends new set of features from ~\cite{bohmeSearchingBetterLife2020}.
Then, we present the performance metric used to compare the forecasting models also used in \cite{robinsonMachineLearningApproach2018}.
Finally, we briefly introduce recurrent neural networks.

\subsection{Data and features sets}

Table~\ref{tab:features} gives an overview of the features used from the data provided by \citeauthor{bohmeSearchingBetterLife2020}~\shortcite{bohmeSearchingBetterLife2020}.
More specifically, we use the following features: both GDP and population size for the origin and destination countries~\cite{worldbankWorldDevelopmentIndicators2020}, the bilateral GTI, migration numbers from the previous year~\cite{oecdInternationalMigrationDatabase2020}, as well as 3 one-hot vectors for encoding the origin, destination and the year.

\paragraph{Google Trends Index features} \label{par:gti}

The Google Trends Index (GTI) is based on the Google Trends data freely accessible at \cite{googleGoogleTrends2020}. The Google Trends tool allows collecting a daily measure of the relative quantities of web search of a precise keyword in a particular region of the world for a specified span of time\footnote{The data can be downloaded from their website, or through an unofficial API~\cite{generalmillsGeneralMillsPytrends2019}.} ~\cite{generalmillsGeneralMillsPytrends2019,googleGoogleTrends2020}. 
To best represent the migration intentions of internet users via online searches, a set of terms related to the theme of migration is selected. 
It is composed of the 67 most semantically related terms to “immigration” and the 67 most semantically related terms to the word “economics” according to the website “Semantic Link”\footnote{\url{https://semantic-link.com/}}. Every term is transcribed in 3 languages with Latin roots: English, Spanish and French to not complicate the extraction too much while covering a maximum of people, that is, about 841 million native speakers~\cite{eberhardEthnologueLanguagesWorld2020}. Tables~\ref{tab:kws_1} and~\ref{tab:kws_2} in the Appendix contain the set of main keywords~\cite{bohmeSearchingBetterLife2020}.

The Google Trends Indexes of a precise keyword for a given country are then calculated by capturing the measures provided by Google Trends for the keyword in the geographical area corresponding to the country for the period spanning from 2004 to 2014\footnote{Google Trends data only starts from 2004 and the migration data stops after 2015.}. Since the values provided by Google are provided as intervals of one month\footnote{This is specific to requests spanning from 2004 to the present.} and are normalized in a range between 0 and 100, the GTI are computed by taking the average of the values for each year to match the migration data.  The indexes, therefore, reflect the variation of the number of searches for the keyword over the years.

The bilateral GTI data is made up of the two different forms of vectors: $GTIbi_{i,j,t}$ and $GTIuni_{i,t} \times GTIdest_{i,j,t}$ for the unilateral and bilateral aspects.
Three different forms of GTI values are then defined: \begin{itemize}     
    \item the vector of unilateral GTI ($GTIuni_{i,t}$) contains the GTI values of the set of keywords for the country of origin $i$ during the year $t$;
    \item the vector of bilateral GTI ($GTIbi_{i,j,t}$) contains GTI values also specific to the country of destination $j$. The values are still captured in the country of origin $i$ during the year $t$ but the related keywords correspond to the combination of the terms with the name of the destination country (e.g., visa Spain, migrant Spain, work Spain);
    \item the destination GTI ($GTIdest_{i,j,t}$) contains only the GTI value of the keyword corresponding to the destination country’s name $j$ (e.g., Spain) for the country $i$ and the year $t$.
\end{itemize}
\paragraph{Migration Data}

The OECD International Migration Database \cite{oecdInternationalMigrationDatabase2020} provides a yearly incoming migratory flow from 101 countries of origin to the 35 countries members of the OECD from the early 1980's until 2015. 
Demographic and economic data about each destination and origin countries have been gathered from the World Development Indicators \cite{worldbankWorldDevelopmentIndicators2020}.


\subsection{Evaluating forecasting models}
\label{subsec:evaluating}

The performance of forecasting models can be evaluated with several metrics. We present below the metrics used in \cite{robinsonMachineLearningApproach2018}. Their formulas are summarized in Table \ref{tab:metrics}. 
\begin{description}     
    \item[Common Part of Commuters ($CPC$):] Its value is 0 when the ground matrix $\mathbf{T}$ and the forecasting matrix $\mathbf{\hat{T}}$ have no entries in common, and 1 when they are identical.
    \item[Mean Absolute Error ($MAE$):] Its value is 0 when the values of both matrices are identical, and arbitrarily positive the worse the forecasting gets.
    \item[Root Mean Square Error ($RMSE$):] Its value is 0 when the values of both matrices are identical, and arbitrarily positive the worse the forecasting gets. The main difference with the MAE is that the RMSE penalizes more strongly the large errors.
    \item[Coefficient of determination (r$^2$):] Its value is 1 when the forecasts perfectly fit the ground truth values, 0 when the forecasts are identical to the expectation of the ground truth values, and arbitrarily negative the worse the fit gets.
    \item[Mean Absolute Error In ($MAE_{in}$):] The MAE on total incoming migrant by destination countries $v_j = \sum_{i = 1}^m T_{i,j}$.
\end{description}

To make fair comparisons, for our experiments, we use these metrics. Remember that the focus of the forecasting is on incoming international migration to OECD countries.

\begin{table*} \caption{The different metrics -- $\mathbf{T}$ is the ground truth value, $\mathbf{\hat{T}}$ is the forecasting matrix, $m$ is the number of origin countries, $n$ is the number of destination countries, $v_j= \sum_{i = 1}^m T_{i,j}$ is the number of incoming migrants for a zone $j$, $\hat{v}_j$ its forecast.}
\label{tab:metrics} \centering \begin{tabular}{@{}ll@{}}     \toprule     \textbf{Metrics} & \textbf{Equations}\\     \midrule
Common Part of Commuters &  \begin{minipage}{.45\textwidth}\begin{flalign}
CPC(\mathbf{T}, \mathbf{\hat{T}}) = \dfrac{2\sum_{i,j=1}^{m, n}\min{(T_{i,j}, \hat{T}_{i,j}})}{\sum_{i,j=1}^{m, n} T_{i,j} + \sum_{i,j=1}^{m, n}\hat{T}_{i,j}}\label{eq:cpc}     \end{flalign}\end{minipage}\\
Mean Absolute Error &  \begin{minipage}{.45\textwidth}\begin{flalign}
MAE(\mathbf{T}, \mathbf{\hat{T}}) = \dfrac{1}{m\cdot n}\sum_{i,j=1}^{m, n}|T_{i,j} - \hat{T}_{i,j}|\label{eq:mae_metric}     \end{flalign}\end{minipage}\\
Root Mean Square Error &  \begin{minipage}{.45\textwidth}\begin{flalign}
RMSE(\mathbf{T}, \mathbf{\hat{T}}) =          \sqrt{\dfrac{1}{m \cdot n}\sum_{i,j=1}^{m,n}(T_{i,j} - \hat{T}_{i,j})^2}\label{eq:rmse_metric}     \end{flalign}\end{minipage}\\
Coefficient of determination &  \begin{minipage}{.45\textwidth}\begin{flalign}         r^2(\mathbf{T}, \mathbf{\hat{T}}) = 1 - \dfrac{\sum_{i,j=1}^{m,n}(T_{i,j} - \hat{T}_{i,j})^2}{\sum_{i,j=1}^{m,n}(T_{i,j} - \overline{T})^2}\label{eq:r_squared}     \end{flalign}\end{minipage}\\
Mean Absolute Error In &  \begin{minipage}{.45\textwidth}\begin{flalign}
MAE_{in}(\mathbf{v}, \mathbf{\hat{v}}) = \dfrac{1}{n}\sum_{j}^{n}|v_{j} - \hat{v}_{j}|\label{eq:mae_in}     \end{flalign}\end{minipage}\\     \bottomrule \end{tabular} \end{table*}

\subsection{Recurrent Neural Networks and Long Short-Term Memory (LSTM)}

Recurrent neural networks (RNN) ~\cite{gravesSupervisedSequenceLabelling2012,goodfellowDeepLearning2016}) are types of artificial neural networks (ANN) architectures particularly well suited to predict time-series or sequential data. It allows sharing features learned across different parts of the sequential data to persist through the network. It is not required to have a fixed set of input vectors. 
Long short-term memory (LSTM)~\cite{hochreiterUntersuchungenDynamischenNeuronalen1991,doyaBifurcationsRecurrentNeural1993,bengioLearningLongtermDependencies1994,gravesSupervisedSequenceLabelling2012} are special architectures of RNN improving their ability to learn properly long-term dependencies by limiting the risk of  vanishing and exploding gradient problems.

LSTM has recently gained momentum for several applications including in forecasting. It performs well in forecasting compared to other ML techniques~\cite{schmidhuberEvolinoHybridNeuroevolution2005,greffLSTMSearchSpace2017,gersLearningForgetContinual2000,yaoDepthGatedLSTM2015,taxPredictiveBusinessProcess2017,liangNeuralNetworkModel2019,gilesNoisyTimeSeries2001,jozefowiczEmpiricalExplorationRecurrent2015}.

\section{Our LSTM approach} 
\label{sec:approach}

Figure~\ref{fig:unfolded_lstm} shows the architecture of our RNN. We use one RNN in charge of forecasting the bilateral flows $\hat{T}_{i,j, t+1}$ with the origin and destination countries, one hot encoded for all pairs $(i,j)$.  
The RNN has a unique LSTM layer. Another approach would have been to use different networks to estimate the flow for each pair of countries. The amount of data to train each would have been very limited, though. We use an additional densely-connected neural network layer on top of the gated-RNN layer, which computes the output scalar value $\hat{T}_{i,j, t+1}$ given hidden vectors provided by the LSTM\footnote{It is common to use an activation function such as softmax with LSTM. However, in this work the output is scalar and we do not use an activation function for the densely connected neural network layer.}.

\subsection{Learning models and hyper-parameter optimization}

To train the ML models we proceed using three sets~\cite{goodfellowDeepLearning2016}: (a) a train set, gathering the input features from 2004 to 2012 (input \_features$_{i,j,04..12}$ $\forall i,j$) and also all the observed migration flows spanning from 2005 to 2013 as output (T$_{i,j,05..13}$ $\forall i,j$) since we forecast next year migration; (b) a validation set, containing input features on the year 2013 (input\_features$_{i,j,13}$ $\forall i,j$) and migration flows of 2014 (T$_{i,j,14}$ $\forall i,j$); and (c) a test set, on the year 2014 (input\_features$_{i,j,14}$ $\forall i,j$) and 2015 (T$_{i,j,15}$ $\forall i,j$). The hyperparameters are optimized accordingly for each model.

The data set is composed of 101 countries of origin, 34 countries of destination, 1997 time-series of length in the range from 2 to 11 for a total of 19 326 observations. Since the validation and the test sets each gather data for one of the 11 years available, each of these sets represents slightly less than 10\% of the whole data.

A simplified version of our LSTM training is presented in Algorithm~\ref{alg:train_model}, while our LSTM evaluation is presented in Algorithm~\ref{alg:validation_test_model}. Notice that the span of years presented in the algorithms corresponds to the one used once the validation is completed, that is, we fit our model on both the training and validation set.

\begin{algorithm}
     \KwData{\textit{model}: LSTM untrained model}
     \KwResult{Model is trained}
     \For{each epoch}{
        \For{each pair i,j of origin-destination countries}{
            \tcc{gradient descent for each batch:}
            \textit{model}.fit(input\_features$_{i,j,04..13}$ , T$_{i,j,05..14}$)
        }
     }
    evaluation(\textit{model}) \tcc{see algorithm 2}
     \caption{Our training algorithm}
     \label{alg:train_model}
\end{algorithm}

\begin{algorithm}
     \KwData{\textit{model}: LSTM trained model }
     \KwResult{Model is evaluated}
     \For{each pair i,j of origin-destination countries}{
        $\hat{T}_{i,j,15}$ $\leftarrow$ \textit{model}.forecast(input\_features$_{i,j,04..14}$) \\
        error $\leftarrow$ compute\_metrics($T_{15}$,$\hat{T}_{15}$)
     }
     \caption{Our evaluation algorithm}
     \label{alg:validation_test_model}
\end{algorithm}

Due to the specificity of LSTM, we fit our LSTM time series by time series\footnote{By time series we mean the sequence of annual migration flows between a pair of origin-destination.}.
Therefore we use a batch of the size corresponding to the number of years present in the series. This implies that the gradient descent is applied and the LSTM's parameters are updated after each propagation of a time series through the LSTM cells (as presented in Figure~\ref{fig:unfolded_lstm}).
Furthermore, the features have been normalized by time series of origin-destination using a min-max scaler~\cite{goodfellowDeepLearning2016}. Our LSTM model uses a bias of 1 for the LSTM forget gate since it has been shown to improve performances drastically~\cite{gersLearningForgetContinual2000,jozefowiczEmpiricalExplorationRecurrent2015}. 

We make use of dropout regularization to reduce the overfitting for both models and ensure a better generalization~\cite{srivastavaDropoutSimpleWay2014,galTheoreticallyGroundedApplication2016,arpitCloserLookMemorization2017}

For the experiments, we have used three different loss functions: (a) Mean Absolute Error ($MAE$); (b) Mean Square Error ($MSE$); and (c) Common Part of Commuters ($CPC$, as described in \cite{robinsonMachineLearningApproach2018}, that is, $loss_{cpc} = 1 - cpc$ with cpc given by Equation \eqref{eq:cpc} ) and adapt them to handle time series.

We optimize the following hyperparameters and present them along with their optimal value: loss function - \textit{MAE} using \textit{Adam} optimizer, number and size of hidden layers - \textit{1} layer of width \textit{50}, number of epochs - \textit{50}, and dropout - \textit{0.15}.

\section{Results and discussion}
\label{sec:results_and_discussion}

\begin{table*}
\caption{Comparison of the 3 models for the specified metrics. The values are shown by pair (train - test). Bold values indicate the best values per column. There are on average 742 migrants  and 46 119 incoming migrants.}
\label{tab:metrics_cmp}
\centering
\begin{tabular}{@{}lrrrrrrrrrrrrrr@{}}
    \toprule
    & \multicolumn{2}{c}{CPC} & & \multicolumn{2}{c}{MAE} & & \multicolumn{2}{c}{RMSE} & & \multicolumn{2}{c}{$r^2$} & & \multicolumn{2}{c}{MAE$_{in}$}\\\cmidrule{2-3} \cmidrule{5-6} \cmidrule{8-9} \cmidrule{11-12} \cmidrule{14-15}
    Models & train & test & & train & test & & train & test & & train & test & & train & test \\\midrule
    Gravity & 0.871 & 0.866 & & 819 & 877 & & 6 100 & 5 239 & & 0.800 & 0.773 & & 24 128 & 28 737 \\
    ANN & 0.931 & 0.834 & & 119 & 306 & & 818 & 1 553 & & 0.975 & 0.921 & & 3 257 & 9 664 \\
    LSTM & \textbf{0.945} & \textbf{0.892} & & \textbf{96} & \textbf{225} & & \textbf{639} & \textbf{1 028} & & \textbf{0.985} & \textbf{0.967} & & \textbf{2 261} & \textbf{4 827} \\
    \bottomrule
\end{tabular}
\end{table*}

We carry out experiments comparing the performance of our LSTM approach with two other models: 
\begin{enumerate}[label=(\alph*)]
    \item the bilateral gravity model estimated through an OLS model as presented in \shortcite{bohmeSearchingBetterLife2020} whose gravity equation is represented below:
    \begin{equation}
    \centering
    \begin{split}
        \log T_{i,j,t+1} =  & \ \beta_1 \ GTIbi_{i,j,t} + \beta_2 \ GTIuni_{i,t} \times GTIdest_{i,j,t} + \\ & \ \beta_3 \ \log GDP_{i,t}  +  \beta_4 \ \log pop_{i,t}+ \beta_5 \ \log GDP_{j,t}+ \\ & \ \beta_6 \ \log pop_{j,t} +  fixed_i + fixed_j + fixed_t + \epsilon_{i, j, t}
    \end{split}    
    \label{eq:ols_model_bi}
    \end{equation}
    With $\epsilon_{i, j, t}$ representing the robust error term;
    \item a deep learning based artificial neural network model (ANN model) as proposed in \shortcite{robinsonMachineLearningApproach2018}. The ANN is composed of densely connected with rectified linear units (ReLU) activation layers. We use the same model for all the forecasts with a time-step of 1 year. This means that the ANN receives as input the set of features $input \ features_{i,j,t}$ described in Table~\ref{tab:features} and outputs the forecasted next-year migration flow $T_{i,j,t+1}$. We optimize the following hyperparameters and present them along with their optimal value: loss function - \textit{MAE} using \textit{Adam} optimizer, number and width of hidden layers - \textit{2} layers of width \textit{200}, training batch size - \textit{32}, number of epochs - \textit{170} and dropout - \textit{0.1}. 
\end{enumerate}


Our source code is available on the following git repository: \url{https://github.com/aia-uclouvain/gti-mig-paper}. It contains the script to extract the Google Trends Index, the Google Colab notebook to build the different models, as well as the data we used. The code is written in python and uses the Keras library, which runs on top of TensorFlow. 

To assess the forecasting power of each model, we use a test set represented by every migration flow taking place in 2015 which represents a bit less than 10\% of the whole data.

Table~\ref{tab:metrics_cmp} shows the results of each model on both the training and test sets for the five metrics described in section \ref{subsec:evaluating}.

Clearly, the ML models perform much better than the ~\citeauthor{bohmeSearchingBetterLife2020}~\shortcite{bohmeSearchingBetterLife2020}’s gravity model. Indeed, with the same data, the ANN is better than the first model in almost every metric while the LSTM model completely outperforms it in all the measures. The ANN model fits very well with the training data but it does not seem to generalize as well as the LSTM model as shown by their performance on the test set. On this data set, the LSTM is the best forecasting model among these three.

Note that, from Table~\ref{tab:metrics_cmp} the RMSE values are always higher than the MAE (between 5 and 7 times larger). We can conclude that the models tend to make a few really large errors. This can be explained by analyzing the data. In the data set, the mean value of migration flows between 2 countries during a year is 742 but the median value is only 17 while the maximum is about 190 000. This indicates that our data set is very sparse: there is a lot of near-zero observations (40\% are below 10) for a very few extremely important ones (less than 2\% reach 10 000). Notice that the mean absolute errors of the different models are very important compared to the mean annual migration flows (742 and 46 119, see Table~\ref{tab:metrics_cmp} caption) but these values are heavily biased by the sparsity of the data and by the large errors made on the really large migration flows (e.g., the USA and Spain).


\begin{figure}
	\centering
    \includegraphics[width=\columnwidth]{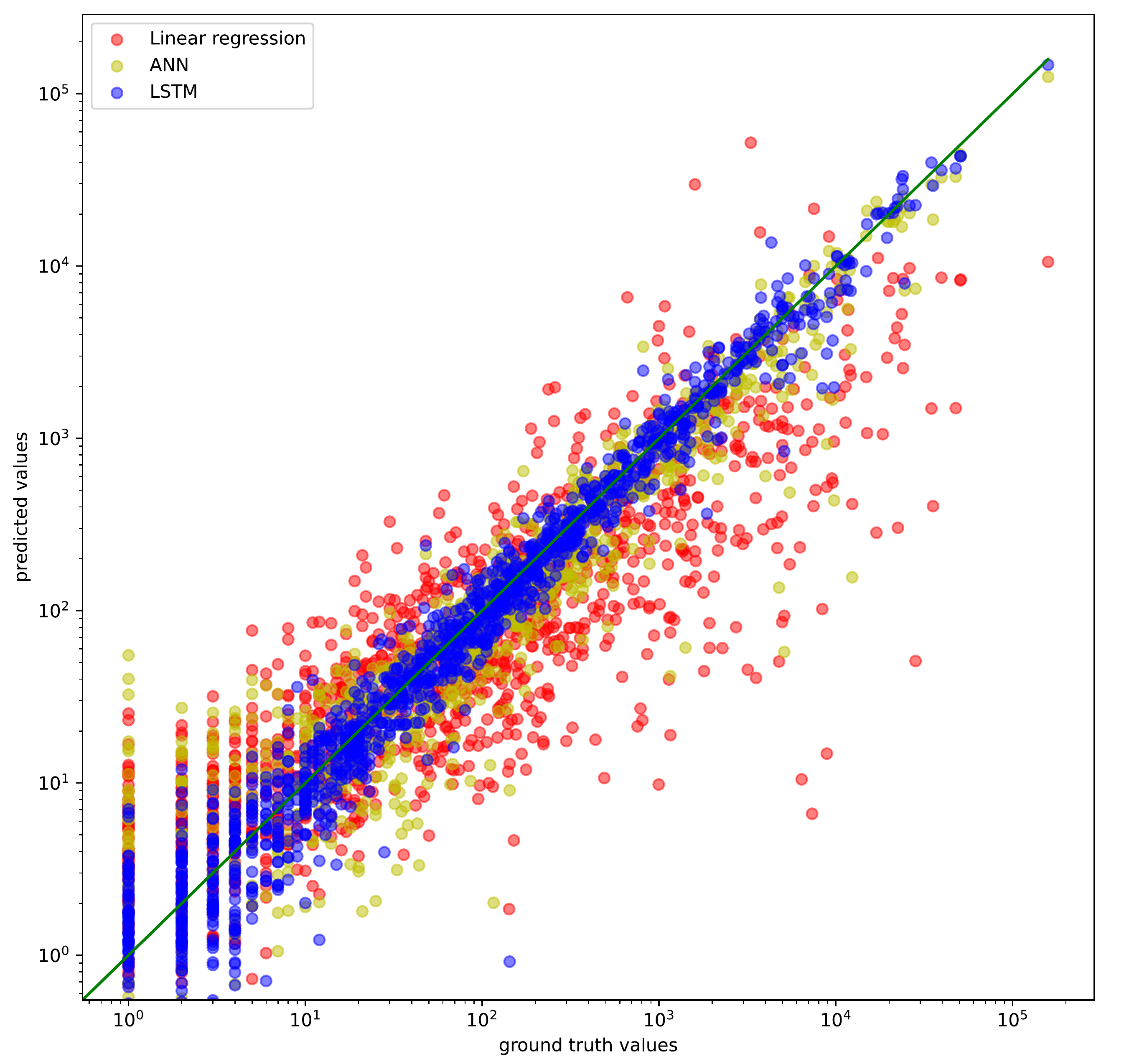}
	\caption{Scatter plot for the 3 models on the test set (year 2015) -- The coefficient of determination for the linear regression is $0.773$, for the ANN $0.921$, and for the LSTM $0.967$ (see the Table~\ref{tab:metrics_cmp} for more details).}
	\label{fig:scatter_all}
\end{figure}

To have a better visualization of the forecasting power of the models, we represent in Figure~\ref{fig:scatter_all} the scatter plot of the 3 models for the test set only. The graph reflects well the sparse nature of the data as shown by the density of points along the x-axis. As expected following the first results, we can observe that the gravity model does not provide very accurate forecasts. The ANN model, on the other hand, shows a stronger tendency to underestimate the ground truth values. Ultimately, the LSTM’s estimations are the ones sticking the most to the actual migration flows which confirm our first assumption.

\begin{figure*}
	\centering
        \includegraphics[width=1.0\textwidth]{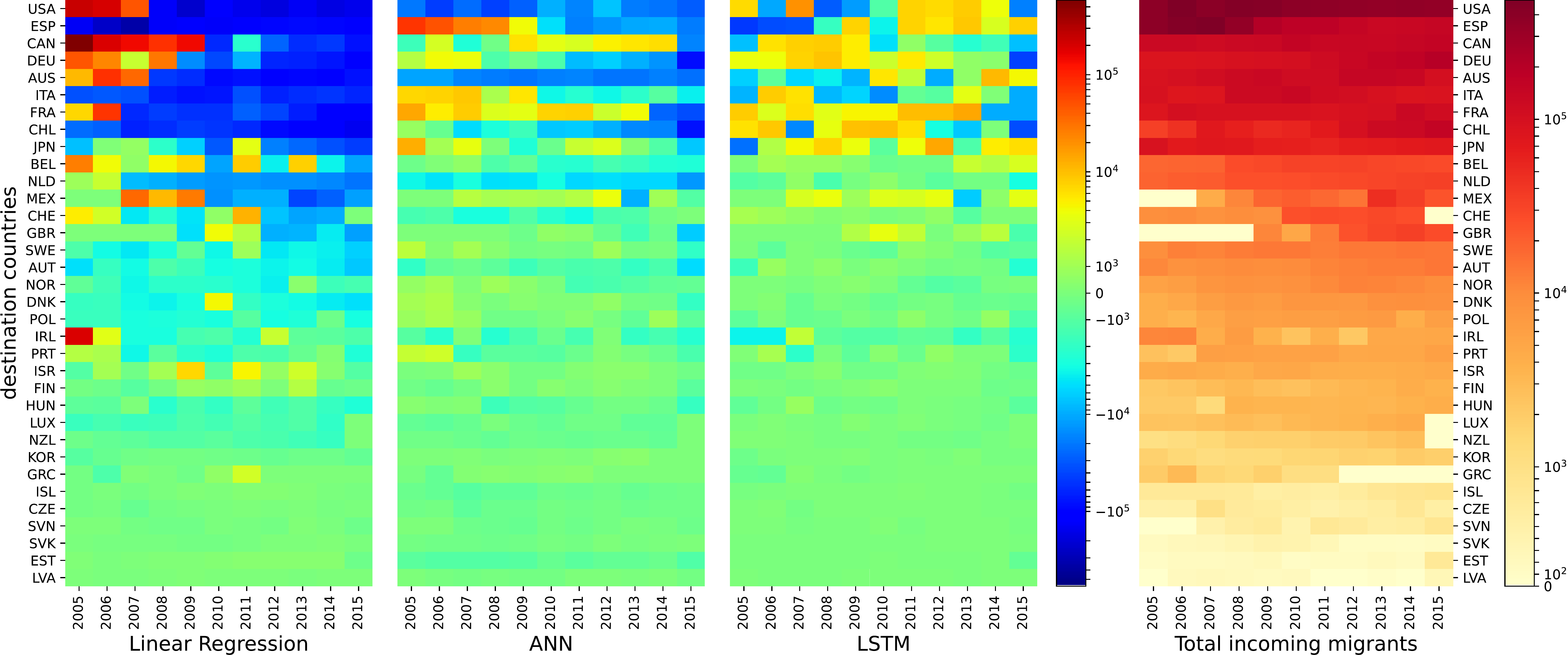}
		\caption{Heat maps of the error on total incoming migrants for 34 OECD countries on the test year (2015) showing how well each model fits the data. From left to right: Gravity Model, ANN Model, and LSTM Model. The rightmost figure is the ground values for the total number of incoming migrants by destination countries. Countries are in descending order of total incoming migrants.}
		\label{fig:dest_all}
\end{figure*}

Finally, Figure~\ref{fig:dest_all} shows the error of the total number of incoming migrants per destination country per year for each model. We can observe that whatever the model, for the majority of countries and years, the estimation error is close to null and that the big errors often appear in the same countries of destination. Knowing that we can see that the heat maps of the ANN model and the linear regression in Figure \ref{fig:dest_all} highlight their tendency to underestimate the migration flows especially for the last year (the test year).

To compare these errors with the actual migration flows, we represent in the rightmost heat map in Figure~\ref{fig:dest_all} the ground truth values of the total number of incoming migrants per destination country and per year in descending order. 
With this figure, we can clearly see that the errors we make are mostly for the countries with important incoming migration flow.

In the case of Spain notice that there has been an important drop in incoming migration flows in 2008 due to the 2007--2008 financial crisis \cite{domingoSistemaMigratorioHispanoAmericano2017}. If we look at the LSTM model in Figure \ref{fig:dest_all}, we largely underestimate the forecasts for Spain from 2005 to 2007. From 2008 and onwards, the forecast errors are comparatively smaller to those before that pivot year. This might indicate a lack of complexity of our model as it does not take into account major past events like a financial crisis.

Table \ref{tab:metrics_cmp}, Figures \ref{fig:scatter_all} and \ref{fig:dest_all} confirm our empirical results that the LSTM approach is able to forecast better than both a gravity model and an ANN model on the data set using different metrics.

\section{Conclusion}
\label{sec:conclusion}
\citeauthor{bohmeSearchingBetterLife2020}~\shortcite{bohmeSearchingBetterLife2020}  have recently demonstrated that including Google Trends data in the set of standard features could improve the migration forecasting models.
In this work, relying exactly on the same data, we improved the quality of the forecast significantly by replacing the gravity model used in by a Long short-term memory (LSTM) artificial recurrent neural network (RNN) architecture. Our experiments also demonstrated that the LSTM was outperforming a standard ANN on this task.

A limitation of our testing procedure is we do not test the model on unknown pairs of origin-destination countries. Instead of splitting the training and the test on the years, we could split them according to the countries such that there is no country overlap. This could help us analyze the universality of our model and whether it generalizes properly on unknown countries or not.

Another possible improvement could be to compute $\hat{T}_{i,j,t+1}$ not from the hidden vector of the last cell, but rather from an interpolation of the $n$ previous cells. It might better reduce the impact of the abnormalities due to a specific year. 

Moreover, the results have shown that the models lack some information to acknowledge important variations of the number of incoming migrants in some destination countries. Adding some factors like the presence of catastrophic events (financial crisis, war, natural disasters, epidemics), unemployment, and the share of internet users could significantly improve our approaches.

In this work, we used categorical labels in form of one-hot vector ($fixed_{i}$, $fixed_{j}$, and $fixed_{t}$). One could use a single one-hot vector $fixed_{i,j}$ containing a pair origin-destination countries, absorbing more complex time-invariant factors like the distance between the 2 countries, and the presence of common language. ML literature has presented different techniques to handle such inputs ~\cite{potdarComparativeStudyCategorical2017,guoEntityEmbeddingsCategorical2016}.

Finally, a drawback of our machine learning approach is that we lose the interpretability of the model and forecasts despite the high interpretability potential of Google search keywords. 

As future work, we would like to apply the latest interpretability techniques (see \cite{molnar2019interpretable}) to better identify the most important features for making high-quality migration forecasts.
This would equip economists, demographers, and experts in migration with new tools to shed light on migration mechanisms.

\label{sec:acks}
\begin{acks}
The authors acknowledge financial support from the UCLouvain ARC convention on "\emph{New approaches to understanding and modelling global migration trends}" (convention 18/23-091).
\end{acks}

\bibliographystyle{ACM-Reference-Format}
\bibliography{refs}


\begin{thebibliography}{38}


\ifx \showCODEN    \undefined \def \showCODEN     #1{\unskip}     \fi
\ifx \showDOI      \undefined \def \showDOI       #1{#1}\fi
\ifx \showISBNx    \undefined \def \showISBNx     #1{\unskip}     \fi
\ifx \showISBNxiii \undefined \def \showISBNxiii  #1{\unskip}     \fi
\ifx \showISSN     \undefined \def \showISSN      #1{\unskip}     \fi
\ifx \showLCCN     \undefined \def \showLCCN      #1{\unskip}     \fi
\ifx \shownote     \undefined \def \shownote      #1{#1}          \fi
\ifx \showarticletitle \undefined \def \showarticletitle #1{#1}   \fi
\ifx \showURL      \undefined \def \showURL       {\relax}        \fi
\providecommand\bibfield[2]{#2}
\providecommand\bibinfo[2]{#2}
\providecommand\natexlab[1]{#1}
\providecommand\showeprint[2][]{arXiv:#2}

\bibitem[\protect\citeauthoryear{Ahmed, Barlacchi, Braghin, Ferretti, Lonij,
  Nair, Novack, Paraszczak, and Toor}{Ahmed et~al\mbox{.}}{2016}]%
        {ahmedMultiScaleApproachDataDriven2016}
\bibfield{author}{\bibinfo{person}{Mohammed~N Ahmed}, \bibinfo{person}{Gianni
  Barlacchi}, \bibinfo{person}{Stefano Braghin}, \bibinfo{person}{Michele
  Ferretti}, \bibinfo{person}{Vincent Lonij}, \bibinfo{person}{Rahul Nair},
  \bibinfo{person}{Rana Novack}, \bibinfo{person}{Jurij Paraszczak}, {and}
  \bibinfo{person}{Andeep~S Toor}.} \bibinfo{year}{2016}\natexlab{}.
\newblock \showarticletitle{A {{Multi}}-{{Scale Approach}} to {{Data}}-{{Driven
  Mass Migration Analysis}}}.
\newblock \bibinfo{journal}{\emph{SoGood@ ECML-PKDD}} (\bibinfo{year}{2016}),
  \bibinfo{pages}{17}.
\newblock


\bibitem[\protect\citeauthoryear{Anderson}{Anderson}{2011}]%
        {andersonGravityModel2011}
\bibfield{author}{\bibinfo{person}{James~E. Anderson}.}
  \bibinfo{year}{2011}\natexlab{}.
\newblock \showarticletitle{The {{Gravity Model}}}.
\newblock \bibinfo{journal}{\emph{Annual Review of Economics}}
  \bibinfo{volume}{3}, \bibinfo{number}{1} (\bibinfo{date}{Sept.}
  \bibinfo{year}{2011}), \bibinfo{pages}{133--160}.
\newblock
\showISSN{1941-1383, 1941-1391}
\urldef\tempurl%
\url{https://doi.org/10.1146/annurev-economics-111809-125114}
\showDOI{\tempurl}


\bibitem[\protect\citeauthoryear{Arpit, Jastrzebski, Ballas, Krueger, Bengio,
  Kanwal, Maharaj, Fischer, Courville, Bengio, and {Lacoste-Julien}}{Arpit
  et~al\mbox{.}}{2017}]%
        {arpitCloserLookMemorization2017}
\bibfield{author}{\bibinfo{person}{Devansh Arpit}, \bibinfo{person}{Stanislaw
  Jastrzebski}, \bibinfo{person}{Nicolas Ballas}, \bibinfo{person}{David
  Krueger}, \bibinfo{person}{Emmanuel Bengio}, \bibinfo{person}{Maxinder~S.
  Kanwal}, \bibinfo{person}{Tegan Maharaj}, \bibinfo{person}{Asja Fischer},
  \bibinfo{person}{Aaron Courville}, \bibinfo{person}{Yoshua Bengio}, {and}
  \bibinfo{person}{Simon {Lacoste-Julien}}.} \bibinfo{year}{2017}\natexlab{}.
\newblock \showarticletitle{A {{Closer Look}} at {{Memorization}} in {{Deep
  Networks}}}.
\newblock \bibinfo{journal}{\emph{arXiv:1706.05394 [cs, stat]}}
  (\bibinfo{date}{July} \bibinfo{year}{2017}).
\newblock
\showeprint[arxiv]{cs, stat/1706.05394}


\bibitem[\protect\citeauthoryear{Askitas and Zimmermann}{Askitas and
  Zimmermann}{2015}]%
        {askitasInternetDataSource2015}
\bibfield{author}{\bibinfo{person}{Nikolaos Askitas} {and}
  \bibinfo{person}{Klaus~F. Zimmermann}.} \bibinfo{year}{2015}\natexlab{}.
\newblock \showarticletitle{The Internet as a Data Source for Advancement in
  Social Sciences}.
\newblock \bibinfo{journal}{\emph{International Journal of Manpower}}
  \bibinfo{volume}{36}, \bibinfo{number}{1} (\bibinfo{year}{2015}),
  \bibinfo{pages}{2--12}.
\newblock


\bibitem[\protect\citeauthoryear{Bengio, Simard, and Frasconi}{Bengio
  et~al\mbox{.}}{1994}]%
        {bengioLearningLongtermDependencies1994}
\bibfield{author}{\bibinfo{person}{Yoshua Bengio}, \bibinfo{person}{Patrice
  Simard}, {and} \bibinfo{person}{Paolo Frasconi}.}
  \bibinfo{year}{1994}\natexlab{}.
\newblock \showarticletitle{Learning Long-Term Dependencies with Gradient
  Descent Is Difficult}.
\newblock \bibinfo{journal}{\emph{IEEE transactions on neural networks}}
  \bibinfo{volume}{5}, \bibinfo{number}{2} (\bibinfo{year}{1994}),
  \bibinfo{pages}{157--166}.
\newblock


\bibitem[\protect\citeauthoryear{B{\"o}hme, Gr{\"o}ger, and
  St{\"o}hr}{B{\"o}hme et~al\mbox{.}}{2020}]%
        {bohmeSearchingBetterLife2020}
\bibfield{author}{\bibinfo{person}{Marcus~H. B{\"o}hme},
  \bibinfo{person}{Andr{\'e} Gr{\"o}ger}, {and} \bibinfo{person}{Tobias
  St{\"o}hr}.} \bibinfo{year}{2020}\natexlab{}.
\newblock \showarticletitle{Searching for a Better Life: {{Predicting}}
  International Migration with Online Search Keywords}.
\newblock \bibinfo{journal}{\emph{Journal of Development Economics}}
  \bibinfo{volume}{142} (\bibinfo{date}{Jan.} \bibinfo{year}{2020}),
  \bibinfo{pages}{102347}.
\newblock
\showISSN{0304-3878}
\urldef\tempurl%
\url{https://doi.org/10.1016/j.jdeveco.2019.04.002}
\showDOI{\tempurl}


\bibitem[\protect\citeauthoryear{Choi and Varian}{Choi and Varian}{2012}]%
        {choiPredictingPresentGoogle2012}
\bibfield{author}{\bibinfo{person}{Hyunyoung Choi} {and} \bibinfo{person}{Hal
  Varian}.} \bibinfo{year}{2012}\natexlab{}.
\newblock \showarticletitle{Predicting the {{Present}} with {{Google Trends}}}.
\newblock \bibinfo{journal}{\emph{Economic Record}} \bibinfo{volume}{88},
  \bibinfo{number}{s1} (\bibinfo{year}{2012}), \bibinfo{pages}{2--9}.
\newblock
\showISSN{1475-4932}
\urldef\tempurl%
\url{https://doi.org/10.1111/j.1475-4932.2012.00809.x}
\showDOI{\tempurl}


\bibitem[\protect\citeauthoryear{Domingo}{Domingo}{2017}]%
        {domingoSistemaMigratorioHispanoAmericano2017}
\bibfield{author}{\bibinfo{person}{Andreu Domingo}.}
  \bibinfo{year}{2017}\natexlab{}.
\newblock \showarticletitle{{El Sistema Migratorio Hispano-Americano del Siglo
  XXI M{\'e}xico y Espa{\~n}a}}.
\newblock \bibinfo{journal}{\emph{Revista de Ciencas y Humanidades -
  Fundaci{\'o}n Ram{\'o}n Areces}} (\bibinfo{date}{Dec.} \bibinfo{year}{2017}).
\newblock


\bibitem[\protect\citeauthoryear{Doya}{Doya}{1993}]%
        {doyaBifurcationsRecurrentNeural1993}
\bibfield{author}{\bibinfo{person}{Kenji Doya}.}
  \bibinfo{year}{1993}\natexlab{}.
\newblock \showarticletitle{Bifurcations of Recurrent Neural Networks in
  Gradient Descent Learning}.
\newblock \bibinfo{journal}{\emph{IEEE Transactions on neural networks}}
  \bibinfo{volume}{1}, \bibinfo{number}{75} (\bibinfo{year}{1993}),
  \bibinfo{pages}{218}.
\newblock


\bibitem[\protect\citeauthoryear{Eberhard, Gary, and Charles}{Eberhard
  et~al\mbox{.}}{2020}]%
        {eberhardEthnologueLanguagesWorld2020}
\bibfield{author}{\bibinfo{person}{David~M. Eberhard},
  \bibinfo{person}{F.~Simons Gary}, {and} \bibinfo{person}{D.~Fennig Charles}.}
  \bibinfo{year}{2020}\natexlab{}.
\newblock \bibinfo{title}{Ethnologue: {{Languages}} of the {{World}}}.
\newblock \bibinfo{howpublished}{https://www.ethnologue.com/}.
\newblock


\bibitem[\protect\citeauthoryear{Gal and Ghahramani}{Gal and
  Ghahramani}{2016}]%
        {galTheoreticallyGroundedApplication2016}
\bibfield{author}{\bibinfo{person}{Yarin Gal} {and} \bibinfo{person}{Zoubin
  Ghahramani}.} \bibinfo{year}{2016}\natexlab{}.
\newblock \showarticletitle{A {{Theoretically Grounded Application}} of
  {{Dropout}} in {{Recurrent Neural Networks}}}.
\newblock \bibinfo{journal}{\emph{arXiv:1512.05287 [stat]}}
  (\bibinfo{date}{Oct.} \bibinfo{year}{2016}).
\newblock
\showeprint[arxiv]{stat/1512.05287}


\bibitem[\protect\citeauthoryear{{General Mills}}{{General Mills}}{2019}]%
        {generalmillsGeneralMillsPytrends2019}
\bibfield{author}{\bibinfo{person}{{General Mills}}.}
  \bibinfo{year}{2019}\natexlab{}.
\newblock \bibinfo{title}{{{GeneralMills}}/Pytrends}.
\newblock \bibinfo{howpublished}{General Mills}.
\newblock


\bibitem[\protect\citeauthoryear{Gers, Schmidhuber, and Cummins}{Gers
  et~al\mbox{.}}{2000}]%
        {gersLearningForgetContinual2000}
\bibfield{author}{\bibinfo{person}{Felix~A. Gers}, \bibinfo{person}{J{\"u}rgen
  Schmidhuber}, {and} \bibinfo{person}{Fred Cummins}.}
  \bibinfo{year}{2000}\natexlab{}.
\newblock \showarticletitle{Learning to {{Forget}}: {{Continual Prediction}}
  with {{LSTM}}}.
\newblock \bibinfo{journal}{\emph{Neural Computation}} \bibinfo{volume}{12},
  \bibinfo{number}{10} (\bibinfo{date}{Oct.} \bibinfo{year}{2000}),
  \bibinfo{pages}{2451--2471}.
\newblock
\showISSN{0899-7667, 1530-888X}
\urldef\tempurl%
\url{https://doi.org/10.1162/089976600300015015}
\showDOI{\tempurl}


\bibitem[\protect\citeauthoryear{Giles}{Giles}{2001}]%
        {gilesNoisyTimeSeries2001}
\bibfield{author}{\bibinfo{person}{C~Lee Giles}.}
  \bibinfo{year}{2001}\natexlab{}.
\newblock \showarticletitle{Noisy {{Time Series Prediction}} Using {{Recurrent
  Neural Networks}} and {{Grammatical Inference}}}.
\newblock \bibinfo{journal}{\emph{Machine learning}} \bibinfo{volume}{44},
  \bibinfo{number}{1-2} (\bibinfo{year}{2001}), \bibinfo{pages}{161--183}.
\newblock


\bibitem[\protect\citeauthoryear{Ginsberg, Mohebbi, Patel, Brammer, Smolinski,
  and Brilliant}{Ginsberg et~al\mbox{.}}{2009}]%
        {google_flue}
\bibfield{author}{\bibinfo{person}{Jeremy Ginsberg}, \bibinfo{person}{Matthew~H
  Mohebbi}, \bibinfo{person}{Rajan~S Patel}, \bibinfo{person}{Lynnette
  Brammer}, \bibinfo{person}{Mark~S Smolinski}, {and} \bibinfo{person}{Larry
  Brilliant}.} \bibinfo{year}{2009}\natexlab{}.
\newblock \showarticletitle{Detecting influenza epidemics using search engine
  query data}.
\newblock \bibinfo{journal}{\emph{Nature}} \bibinfo{volume}{457},
  \bibinfo{number}{7232} (\bibinfo{year}{2009}), \bibinfo{pages}{1012--1014}.
\newblock


\bibitem[\protect\citeauthoryear{Goodfellow, Bengio, and Courville}{Goodfellow
  et~al\mbox{.}}{2016}]%
        {goodfellowDeepLearning2016}
\bibfield{author}{\bibinfo{person}{Ian Goodfellow}, \bibinfo{person}{Yoshua
  Bengio}, {and} \bibinfo{person}{Aaron Courville}.}
  \bibinfo{year}{2016}\natexlab{}.
\newblock \bibinfo{booktitle}{\emph{Deep Learning}}.
\newblock \bibinfo{publisher}{{MIT press}}.
\newblock


\bibitem[\protect\citeauthoryear{{Google}}{{Google}}{2020}]%
        {googleGoogleTrends2020}
\bibfield{author}{\bibinfo{person}{{Google}}.} \bibinfo{year}{2020}\natexlab{}.
\newblock \bibinfo{title}{Google {{Trends}}}.
\newblock \bibinfo{howpublished}{https://www.google.com/trends}.
\newblock


\bibitem[\protect\citeauthoryear{Graves}{Graves}{2012}]%
        {gravesSupervisedSequenceLabelling2012}
\bibfield{author}{\bibinfo{person}{Alex Graves}.}
  \bibinfo{year}{2012}\natexlab{}.
\newblock \bibinfo{booktitle}{\emph{Supervised {{Sequence Labelling}} with
  {{Recurrent Neural Networks}}}}. \bibinfo{series}{Studies in {{Computational
  Intelligence}}}, Vol.~\bibinfo{volume}{385}.
\newblock \bibinfo{publisher}{{Springer Berlin Heidelberg}},
  \bibinfo{address}{{Berlin, Heidelberg}}.
\newblock
\showISBNx{978-3-642-24797-2}
\urldef\tempurl%
\url{https://doi.org/10.1007/978-3-642-24797-2}
\showDOI{\tempurl}


\bibitem[\protect\citeauthoryear{Greff, Srivastava, Koutn{\'i}k, Steunebrink,
  and Schmidhuber}{Greff et~al\mbox{.}}{2017}]%
        {greffLSTMSearchSpace2017}
\bibfield{author}{\bibinfo{person}{Klaus Greff}, \bibinfo{person}{Rupesh~Kumar
  Srivastava}, \bibinfo{person}{Jan Koutn{\'i}k}, \bibinfo{person}{Bas~R.
  Steunebrink}, {and} \bibinfo{person}{J{\"u}rgen Schmidhuber}.}
  \bibinfo{year}{2017}\natexlab{}.
\newblock \showarticletitle{{{LSTM}}: {{A Search Space Odyssey}}}.
\newblock \bibinfo{journal}{\emph{IEEE Transactions on Neural Networks and
  Learning Systems}} \bibinfo{volume}{28}, \bibinfo{number}{10}
  (\bibinfo{date}{Oct.} \bibinfo{year}{2017}), \bibinfo{pages}{2222--2232}.
\newblock
\showISSN{2162-237X, 2162-2388}
\urldef\tempurl%
\url{https://doi.org/10.1109/TNNLS.2016.2582924}
\showDOI{\tempurl}
\showeprint[arxiv]{1503.04069}


\bibitem[\protect\citeauthoryear{Guo and Berkhahn}{Guo and Berkhahn}{2016}]%
        {guoEntityEmbeddingsCategorical2016}
\bibfield{author}{\bibinfo{person}{Cheng Guo} {and} \bibinfo{person}{Felix
  Berkhahn}.} \bibinfo{year}{2016}\natexlab{}.
\newblock \showarticletitle{Entity {{Embeddings}} of {{Categorical
  Variables}}}.
\newblock \bibinfo{journal}{\emph{arXiv:1604.06737 [cs]}}
  (\bibinfo{date}{April} \bibinfo{year}{2016}).
\newblock
\showeprint[arxiv]{cs/1604.06737}


\bibitem[\protect\citeauthoryear{Hochreiter}{Hochreiter}{1991}]%
        {hochreiterUntersuchungenDynamischenNeuronalen1991}
\bibfield{author}{\bibinfo{person}{Sepp Hochreiter}.}
  \bibinfo{year}{1991}\natexlab{}.
\newblock \showarticletitle{Untersuchungen Zu Dynamischen Neuronalen
  {{Netzen}}}.
\newblock \bibinfo{journal}{\emph{Diploma, Technische Universit{\"a}t
  M{\"u}nchen}} \bibinfo{volume}{91}, \bibinfo{number}{1}
  (\bibinfo{year}{1991}).
\newblock


\bibitem[\protect\citeauthoryear{Hochreiter and Schmidhuber}{Hochreiter and
  Schmidhuber}{1997}]%
        {hochreiterLongShorttermMemory1997}
\bibfield{author}{\bibinfo{person}{Sepp Hochreiter} {and}
  \bibinfo{person}{J{\"u}rgen Schmidhuber}.} \bibinfo{year}{1997}\natexlab{}.
\newblock \showarticletitle{Long Short-Term Memory}.
\newblock \bibinfo{journal}{\emph{Neural computation}} \bibinfo{volume}{9},
  \bibinfo{number}{8} (\bibinfo{year}{1997}), \bibinfo{pages}{1735--1780}.
\newblock


\bibitem[\protect\citeauthoryear{Jozefowicz, Zaremba, and Sutskever}{Jozefowicz
  et~al\mbox{.}}{2015}]%
        {jozefowiczEmpiricalExplorationRecurrent2015}
\bibfield{author}{\bibinfo{person}{Rafal Jozefowicz}, \bibinfo{person}{Wojciech
  Zaremba}, {and} \bibinfo{person}{Ilya Sutskever}.}
  \bibinfo{year}{2015}\natexlab{}.
\newblock \showarticletitle{An {{Empirical Exploration}} of {{Recurrent Network
  Architectures}}}. In \bibinfo{booktitle}{\emph{International Conference on
  Machine Learning}}. \bibinfo{pages}{2342--2350}.
\newblock


\bibitem[\protect\citeauthoryear{Letouz{\'e}, Purser, Rodr{\'i}guez, and
  Cummins}{Letouz{\'e} et~al\mbox{.}}{2009}]%
        {letouzeRevisitingMigrationdevelopmentNexus2009}
\bibfield{author}{\bibinfo{person}{Emmanuel Letouz{\'e}}, \bibinfo{person}{Mark
  Purser}, \bibinfo{person}{Francisco Rodr{\'i}guez}, {and}
  \bibinfo{person}{Matthew Cummins}.} \bibinfo{year}{2009}\natexlab{}.
\newblock \showarticletitle{Revisiting the Migration-Development Nexus: {{A}}
  Gravity Model Approach}.
\newblock \bibinfo{journal}{\emph{Human Development Research Paper 44}}
  (\bibinfo{year}{2009}).
\newblock


\bibitem[\protect\citeauthoryear{Liang, Zhang, and Wang}{Liang
  et~al\mbox{.}}{2019}]%
        {liangNeuralNetworkModel2019}
\bibfield{author}{\bibinfo{person}{Hao Liang}, \bibinfo{person}{Meng Zhang},
  {and} \bibinfo{person}{Hailan Wang}.} \bibinfo{year}{2019}\natexlab{}.
\newblock \showarticletitle{A {{Neural Network Model}} for {{Wildfire Scale
  Prediction Using Meteorological Factors}}}.
\newblock \bibinfo{journal}{\emph{IEEE Access}}  \bibinfo{volume}{7}
  (\bibinfo{year}{2019}), \bibinfo{pages}{176746--176755}.
\newblock
\showISSN{2169-3536}
\urldef\tempurl%
\url{https://doi.org/10.1109/ACCESS.2019.2957837}
\showDOI{\tempurl}


\bibitem[\protect\citeauthoryear{Masucci, Serras, Johansson, and Batty}{Masucci
  et~al\mbox{.}}{2013}]%
        {masucciGravityRadiationModels2013}
\bibfield{author}{\bibinfo{person}{A.~Paolo Masucci}, \bibinfo{person}{Joan
  Serras}, \bibinfo{person}{Anders Johansson}, {and} \bibinfo{person}{Michael
  Batty}.} \bibinfo{year}{2013}\natexlab{}.
\newblock \showarticletitle{Gravity versus Radiation Models: {{On}} the
  Importance of Scale and Heterogeneity in Commuting Flows}.
\newblock \bibinfo{journal}{\emph{Physical Review E}} \bibinfo{volume}{88},
  \bibinfo{number}{2} (\bibinfo{year}{2013}), \bibinfo{pages}{022812}.
\newblock


\bibitem[\protect\citeauthoryear{Molnar}{Molnar}{2019}]%
        {molnar2019interpretable}
\bibfield{author}{\bibinfo{person}{Christoph Molnar}.}
  \bibinfo{year}{2019}\natexlab{}.
\newblock \bibinfo{booktitle}{\emph{Interpretable machine learning}}.
\newblock \bibinfo{publisher}{Lulu. com}.
\newblock


\bibitem[\protect\citeauthoryear{{OECD}}{{OECD}}{2020}]%
        {oecdInternationalMigrationDatabase2020}
\bibfield{author}{\bibinfo{person}{{OECD}}.} \bibinfo{year}{2020}\natexlab{}.
\newblock \bibinfo{title}{International Migration Database}.
\newblock
  \bibinfo{howpublished}{https://www.oecd-ilibrary.org/content/data/data-00342-en}.
\newblock


\bibitem[\protect\citeauthoryear{Poot, Alimi, Cameron, and Mar{\'e}}{Poot
  et~al\mbox{.}}{2016}]%
        {pootGravityModelMigration2016b}
\bibfield{author}{\bibinfo{person}{Jacques Poot}, \bibinfo{person}{Omoniyi
  Alimi}, \bibinfo{person}{Michael~P Cameron}, {and} \bibinfo{person}{David~C
  Mar{\'e}}.} \bibinfo{year}{2016}\natexlab{}.
\newblock \showarticletitle{The {{Gravity Model}} of {{Migration}}: {{The
  Successful Comeback}} of an {{Ageing Superstar}} in {{Regional Science}}}.
\newblock  (\bibinfo{year}{2016}), \bibinfo{pages}{27}.
\newblock


\bibitem[\protect\citeauthoryear{Potdar, S., and D.}{Potdar
  et~al\mbox{.}}{2017}]%
        {potdarComparativeStudyCategorical2017}
\bibfield{author}{\bibinfo{person}{Kedar Potdar}, \bibinfo{person}{Taher S.},
  {and} \bibinfo{person}{Chinmay D.}} \bibinfo{year}{2017}\natexlab{}.
\newblock \showarticletitle{A {{Comparative Study}} of {{Categorical Variable
  Encoding Techniques}} for {{Neural Network Classifiers}}}.
\newblock \bibinfo{journal}{\emph{International Journal of Computer
  Applications}} \bibinfo{volume}{175}, \bibinfo{number}{4}
  (\bibinfo{date}{Oct.} \bibinfo{year}{2017}), \bibinfo{pages}{7--9}.
\newblock
\showISSN{09758887}
\urldef\tempurl%
\url{https://doi.org/10.5120/ijca2017915495}
\showDOI{\tempurl}


\bibitem[\protect\citeauthoryear{Robinson and Dilkina}{Robinson and
  Dilkina}{2018}]%
        {robinsonMachineLearningApproach2018}
\bibfield{author}{\bibinfo{person}{Caleb Robinson} {and}
  \bibinfo{person}{Bistra Dilkina}.} \bibinfo{year}{2018}\natexlab{}.
\newblock \showarticletitle{A Machine Learning Approach to Modeling Human
  Migration}. In \bibinfo{booktitle}{\emph{Proceedings of the 1st {{ACM SIGCAS
  Conference}} on {{Computing}} and {{Sustainable Societies}}}}.
  \bibinfo{pages}{1--8}.
\newblock


\bibitem[\protect\citeauthoryear{Schmidhuber, Wierstra, and Gomez}{Schmidhuber
  et~al\mbox{.}}{2005}]%
        {schmidhuberEvolinoHybridNeuroevolution2005}
\bibfield{author}{\bibinfo{person}{J{\"u}rgen Schmidhuber},
  \bibinfo{person}{Daan Wierstra}, {and} \bibinfo{person}{Faustino~J. Gomez}.}
  \bibinfo{year}{2005}\natexlab{}.
\newblock \showarticletitle{Evolino: {{Hybrid}} Neuroevolution/Optimal Linear
  Search for Sequence Prediction}. In \bibinfo{booktitle}{\emph{Proceedings of
  the 19th {{International Joint Conferenceon Artificial Intelligence}}
  ({{IJCAI}})}}.
\newblock


\bibitem[\protect\citeauthoryear{Simini, Gonz{\'a}lez, Maritan, and
  Barab{\'a}si}{Simini et~al\mbox{.}}{2012}]%
        {siminiUniversalModelMobility2012}
\bibfield{author}{\bibinfo{person}{Filippo Simini}, \bibinfo{person}{Marta~C.
  Gonz{\'a}lez}, \bibinfo{person}{Amos Maritan}, {and}
  \bibinfo{person}{Albert-L{\'a}szl{\'o} Barab{\'a}si}.}
  \bibinfo{year}{2012}\natexlab{}.
\newblock \showarticletitle{A Universal Model for Mobility and Migration
  Patterns}.
\newblock \bibinfo{journal}{\emph{Nature}} \bibinfo{volume}{484},
  \bibinfo{number}{7392} (\bibinfo{year}{2012}), \bibinfo{pages}{96--100}.
\newblock


\bibitem[\protect\citeauthoryear{Srivastava, Hinton, Krizhevsky, Sutskever, and
  Salakhutdinov}{Srivastava et~al\mbox{.}}{2014}]%
        {srivastavaDropoutSimpleWay2014}
\bibfield{author}{\bibinfo{person}{Nitish Srivastava},
  \bibinfo{person}{Geoffrey Hinton}, \bibinfo{person}{Alex Krizhevsky},
  \bibinfo{person}{Ilya Sutskever}, {and} \bibinfo{person}{Ruslan
  Salakhutdinov}.} \bibinfo{year}{2014}\natexlab{}.
\newblock \showarticletitle{Dropout: A Simple Way to Prevent Neural Networks
  from Overfitting}.
\newblock \bibinfo{journal}{\emph{The Journal of Machine Learning Research}}
  \bibinfo{volume}{15}, \bibinfo{number}{1} (\bibinfo{date}{Jan.}
  \bibinfo{year}{2014}), \bibinfo{pages}{1929--1958}.
\newblock
\showISSN{1532-4435}


\bibitem[\protect\citeauthoryear{Tax, Verenich, La~Rosa, and Dumas}{Tax
  et~al\mbox{.}}{2017}]%
        {taxPredictiveBusinessProcess2017}
\bibfield{author}{\bibinfo{person}{Niek Tax}, \bibinfo{person}{Ilya Verenich},
  \bibinfo{person}{Marcello La~Rosa}, {and} \bibinfo{person}{Marlon Dumas}.}
  \bibinfo{year}{2017}\natexlab{}.
\newblock \showarticletitle{Predictive {{Business Process Monitoring}} with
  {{LSTM Neural Networks}}}.
\newblock \bibinfo{journal}{\emph{arXiv:1612.02130 [cs, stat]}}
  \bibinfo{volume}{10253} (\bibinfo{year}{2017}), \bibinfo{pages}{477--492}.
\newblock
\urldef\tempurl%
\url{https://doi.org/10.1007/978-3-319-59536-8_30}
\showDOI{\tempurl}
\showeprint[arxiv]{cs, stat/1612.02130}


\bibitem[\protect\citeauthoryear{UN~DESA}{UN~DESA}{2019}]%
        {undesaInternationalMigrantStock2019}
\bibfield{author}{\bibinfo{person}{United Nations Department of Economic {and}
  Social Affairs Population~Division UN~DESA}.}
  \bibinfo{year}{2019}\natexlab{}.
\newblock \bibinfo{title}{International {{Migrant Stock}} 2019.}
\newblock
  \bibinfo{howpublished}{https://www.un.org/en/development/desa/population/index.asp}.
\newblock


\bibitem[\protect\citeauthoryear{{World Bank}}{{World Bank}}{2020}]%
        {worldbankWorldDevelopmentIndicators2020}
\bibfield{author}{\bibinfo{person}{{World Bank}}.}
  \bibinfo{year}{2020}\natexlab{}.
\newblock \bibinfo{title}{World {{Development Indicators}}}.
\newblock
  \bibinfo{howpublished}{https://datacatalog.worldbank.org/dataset/world-development-indicators}.
\newblock


\bibitem[\protect\citeauthoryear{Yao, Cohn, Vylomova, Duh, and Dyer}{Yao
  et~al\mbox{.}}{2015}]%
        {yaoDepthGatedLSTM2015}
\bibfield{author}{\bibinfo{person}{Kaisheng Yao}, \bibinfo{person}{Trevor
  Cohn}, \bibinfo{person}{Katerina Vylomova}, \bibinfo{person}{Kevin Duh},
  {and} \bibinfo{person}{Chris Dyer}.} \bibinfo{year}{2015}\natexlab{}.
\newblock \showarticletitle{Depth-{{Gated LSTM}}}.
\newblock \bibinfo{journal}{\emph{arXiv:1508.03790 [cs]}} (\bibinfo{date}{Aug.}
  \bibinfo{year}{2015}).
\newblock
\showeprint[arxiv]{cs/1508.03790}


\end{thebibliography}

\appendix
\label{appendix}
\section{Used Keywords}

Tables \ref{tab:kws_1} and \ref{tab:kws_2} contain the set of main keywords: "\emph{For GTI data retrieval, both singular and plural as well as male and female forms of these keywords are used where applicable. In the English language, both British and American English spelling is used. All French and Spanish keywords were included with and without accents}" \cite[Table 1]{bohmeSearchingBetterLife2020}.


\begin{table}[H]
    \centering
    \caption{List of main keywords -- First part \protect\cite[Table 1]{bohmeSearchingBetterLife2020}.}
    \label{tab:kws_1}
    \begin{tabular}{@{}lll@{}}
        \toprule
        \textbf{English} & \textbf{French} & \textbf{Spanish}  \\ \midrule
        applicant & candidat & solicitante \\
        arrival & arrivee & llegada \\
        asylum & asile & asilo \\
        benefit & allocation sociale & beneficio \\
        border control & controle frontiere & control frontera \\
        business & entreprise & negocio \\
        citizenship & citoyennete & ciudadania \\
        compensation & compensation & compensacion \\
        consulate & consulat & consulado \\
        contract & contrat & contrato \\
        customs & douane & aduana \\
        deportation & expulsion & deportacion \\
        diaspora & diaspora & diaspora \\
        discriminate & discriminer & discriminar \\
        earning & revenu & ganancia \\
        economy & economie & economia \\
        embassy & ambassade & embajada \\
        emigrant & emigre & emigrante \\
        emigrate & emigrer & emigrar \\
        emigration & emigration & emigracion \\
        employer & employer & empleador \\
        employment & emploi & empleo \\
        foreigner & etranger & extranjero \\
        \bottomrule
    \end{tabular}
\end{table}

\begin{table}[H]
    \centering
    \caption{List of main keywords -- Second part \protect\cite[Table 1]{bohmeSearchingBetterLife2020}.}
    \label{tab:kws_2}
    \begin{tabular}{@{}lll@{}}
        \toprule
        \textbf{English} & \textbf{French} & \textbf{Spanish}  \\ \midrule
        GDP & PIB & PIB \\
        hiring & embauche & contratacion \\
        illegal & illegal & ilegal \\
        immigrant & immigre & inmigrante \\
        immigrate & immigrer & inmigrar \\
        immigration & immigration & inmigracion \\
        income & revenu & ingreso \\
        inflation & inflation & inflacion \\
        internship & stage & pasantia \\
        job & emploi & trabajo \\
        labor & travail & mano de obra \\
        layoff & licenciement & despido \\
        legalization & regularisation & legalizacion \\
        migrant & migrant & migrante \\
        migrate & migrer & migrar \\
        migration & migration & migracion \\
        minimum & minimum & minimo \\
        nationality & nationalite & nacionalidad \\
        naturalization & naturalisation & naturalizacion \\
        passport & passeport & pasaporte \\
        payroll & paie & nomina \\
        pension & retraite & pension \\
        quota & quota & cuota \\
        recession & recession & recesion \\
        recruitment & recrutement & reclutamiento \\
        refugee & refugie & refugiado \\
        remuneration & remuneration & remuneracion \\
        required documents & documents requis & documentos requisito \\
        salary & salaire & sueldo \\
        Schengen & Schengen & Schengen \\
        smuggler & trafiquant & traficante \\
        smuggling & trafic & contrabando \\
        tax & tax & impuesto \\
        tourist & touriste & turista \\
        unauthorized & non autorisee & no autorizado \\
        undocumented & sans papiers & indocumentado \\
        unemployment & chomage & desempleo \\
        union & syndicat & sindicato \\
        unskilled & non qualifies & no capacitado \\
        vacancy & poste vacante & vacante \\
        visa & visa & visa \\
        waiver & exemption & exencion \\
        wage & salaire & salario \\
        welfare & aide sociale & asistencia social \\
        \bottomrule
    \end{tabular}
\end{table}

\end{document}